\documentclass[10pt,twocolumn,letterpaper]{article}

\usepackage{times}
\usepackage{epsfig}
\usepackage{graphicx}
\usepackage{amsmath}
\usepackage{amssymb}
\usepackage{colortbl}

\usepackage{rotating}
\usepackage[font={scriptsize}]{caption}
\usepackage{multirow}

\usepackage[breaklinks=true,bookmarks=false]{hyperref}

\date{\vspace{-5ex}}

\newcommand{\dquote}[1]{``#1''}
\newcommand{\etal}{\textit{et al.}~}

\newcommand\Tstrut{\rule{0pt}{2.3ex}}         

\begin{document}

\title{Segmentation Free Object Discovery in Video}


\author{Giovanni Cuffaro, Federico Becattini, Claudio Baecchi, Lorenzo Seidenari, Alberto Del Bimbo\\
University of Florence\\
{\tt\small \{name.surname\}@unifi.it}}

\maketitle

\begin{abstract}\textit{
   In this paper we present a simple yet effective approach to extend without supervision any object proposal from static images to videos. Unlike previous methods, these spatio-temporal proposals, to which we refer as \dquote{tracks}, are generated relying on little or no visual content by only exploiting bounding boxes spatial correlations through time. The tracks that we obtain are likely to represent objects and are a general-purpose tool to represent meaningful video content for a wide variety of tasks. For unannotated videos, tracks can be used to discover content without any supervision. As further contribution we also propose a novel and dataset-independent method to evaluate a generic object proposal based on the entropy of a classifier output response.
   We experiment on two competitive datasets, namely YouTube Objects \cite{prest2012learning} and ILSVRC-2015 VID \cite{russakovsky2015imagenet}.
}\end{abstract}

\section{Introduction} \label{intro}
Image and video analysis can be considered similar on many levels, but whereas new algorithms are continuously raising the bar for static image tasks, advancements on videos seem to be slower and hard going. What makes video comprehension more difficult is mainly the huge amount of data that has to be processed and the need to model an additional dimension: time.

We believe that focusing on relevant regions of videos, such as objects, will reduce the complexity of the problem and ease learning for models like Deep Networks. The same concept has been successfully applied to images using object proposals, which analyse low level properties, such as edges, to find regions that are likely to contain salient objects. Advantages are twofold, first the search space is considerably reduced, second, as a consequence, the number of false positives generated by classifiers is lowered.

In this work we propose a technique to include time into a generic object proposal, by exploiting the weak supervision provided by time itself to match spatial proposals between adjacent frames.
This results in spatio-temporal tracks that represent salient objects in the video and can therefore be used instead of the whole sequence.
To the best of our knowledge we are the first to adopt a fully unsupervised matching strategy that only relies on bounding box coordinates without any semantic content or visual descriptor apart from optical flow. We also introduce a novel dataset-independent proposal evaluation method based on the entropy of classifier scores.



\begin{figure*}
\centering
  \includegraphics[width=.9\textwidth]{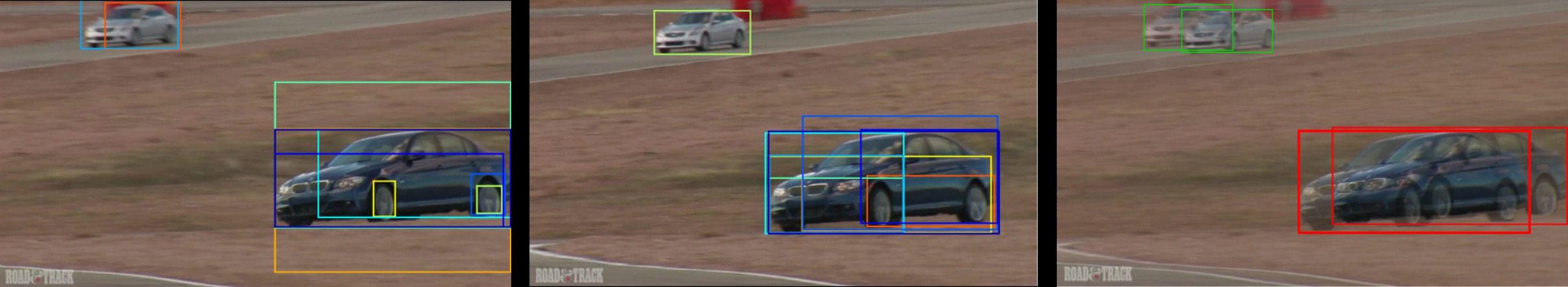}
  \caption{Example of frame matching; matched boxes are inserted into their respective track. \textit{(left)} reference frame where the top 10 proposals extracted with EdgeBoxes are shown; \textit{(center)} following frame with top 10 EdgeBoxes proposals; \textit{(right)} 2 matched proposals between the two frames; these will be part of two different tracks}
  \label{fig:matchingexample}
\end{figure*}


\section{Related Work}

Object proposals \cite{hosang2015} provide a relatively small set of bounding boxes likely to contain salient regions in images, based on some \textit{objectness} measure. Different proposals, such as EdgeBoxes \cite{zitnick2014edge}, are commonly used in image related tasks to reduce the number of candidate regions to evaluate.
Recently, there have been some attempts to adapt the paradigm of object proposals to videos to solve specific tasks, by generating consistent spatio-temporal volumes.
In \cite{prest2012learning} motion segmentation is exploited to extract a single spatio-temporal tube for video, in order to perform video classification. The task of object discovery is tackled in \cite{stretcu2015multiple} by generating a set of boxes using a foreground estimation method and matching them across frames using both geometric and appearance terms. Kwak \etal \cite{kwak2015unsupervised} combine a discovery step matching similar regions in different frames and a tracking step to obtain temporal proposals. In \cite{oneata2014spatio} a classifier is learnt to guide a super-voxel merging process for obtaining object proposals. Temporal proposals have been exploited to segment objects in videos in \cite{xiao2016track} by discovering easy instances and propagating the tube to adjacent frames. Other methods to generate salient tubes have been proposed for action localization in \cite{yu2015fast} using human and motion detection.

Differently from the above approaches we do not rely on segmentation, which is a time-consuming task especially for videos. Our method is simply based on the response of a frame-wise proposal method. The weak supervision obtained from the temporal consistency of the video is exploited to generate tracks. Our method aims at generating few, highly precise, tracks containing objects in the video.



\section{Video Temporal Proposals} \label{tempprop}
In this section we introduce the concept of \dquote{track}, describing in details how these are generated from a set of bounding boxes extracted by an object proposal in the video.

Given a video $V$, for each frame $f_i$ we extract a set $B_i$ of bounding boxes $b_i^k$ using an object proposal. We propose a method to match boxes that exhibit a temporal consistency in consecutive frames through the video, yielding to a set $T$ of tracks $t_j$. A track is defined as a succession of bounding boxes $b_i^k$ for which the intersection over union (IoU) between two boxes $b_i^m$ (belonging to frame $f_i$) and $b_{i+1}^n$ (belonging to frame $f_{i+1}$) is above a defined threshold $\theta_\tau$.

Starting from the first frame, each time a match is found, the corresponding bounding box is added to the end of the track and becomes the reference box for the following frame. If no match is found the last box of the track is compared with the following frames until a good match is obtained. An example of matching is shown in Fig.~\ref{fig:matchingexample}.

When one or more consecutive matches are not found, tracks become fragmented, i.e. there are frames for which a track is active but there is no bounding box.
This is usually due to a lack of good bounding boxes for that frame, occlusion or appearance changes of the object.
It is thus necessary to avoid matching boxes in frames too far apart that therefore do not represent the same content, but at the same time we want to be able to tolerate some missing boxes without prematurely terminating the track.

To this end we introduce a Time to Live counter (TTL) $\tau$ for each track. We define $\tau_i(t_j)$ as the number of frames, at frame $i$, that the method can still wait before considering the track $t_j$ terminated. TTL starts from an initial value $\gamma$; each time a box can not be matched in a consecutive frame the TTL is decremented, otherwise is incremented (up to $\gamma$). More formally, given a track $t_j$ and its last bounding box $b_i^m$ we increment or decrement its TTL as follows:

\begin{equation}
\tau_{i+1}(t_j) =
\begin{cases}
  \tau_{i}(t_j)+1, & \text{if}\ \exists\ n : \text{IoU}(b_i^m, b_{i+1}^n) > \theta_\tau \\
  \tau_{i}(t_j)-1, & \text{otherwise}
\end{cases}
\end{equation}

When the TTL for a track reaches $0$, the track is considered terminated. Missing frames caused by track fragmentation are linearly interpolated using the positions of the previous and following bounding boxes in the track.

\paragraph{Proposal Motion Compensation}
Proposals around objects in consecutive frames are usually unaligned due to movements of the object or the camera. This causes the IoU score to decrease even if the matching is good.
We work around this problem by registering the boxes with optical flow before computing the IoU. The registration is performed on the last box of each track, by computing the mean offsets along the $x$ and $y$ axes inside the boxes. Shifted boxes are only used for matching and tracks consist only of unaltered boxes.


%
%

\paragraph{Temporal NMS}
As in the spatial case, temporal proposals also suffer of high redundancy. To reduce this effect we extend spatial non-maximal suppression to time, defining a temporal NMS  where instead of computing IoU on areas it is computed over volumes (vIoU). If $\alpha_j^k$ is the area of the $k$-th bounding box in track $t_j$, then the volume $\upsilon_j$ of the track is calculated as $\upsilon_j = \sum_{k~=~0}^{K}\alpha_j^k$
where $K$ is the length of the track. Then, vIoU is defined as:

\begin{equation}
	\text{vIoU}(t_j, t_k) = \frac{\upsilon_j~\cap~\upsilon_k}{\upsilon_j~\cup~\upsilon_k}
\end{equation}

Using vIoU we apply the standard NMS.

\paragraph{Proposal Suppression}
Once all the tracks are computed for a given video, we apply a post-processing to remove the ones that are unlikely to represent an object. To this end we remove those tracks which have a length smaller than a value $l$. In this way we exclude very short tracks that are likely to be composed by background boxes that happen to have a high IoU.

Another problem is posed by logos and writings impressed on the video. In fact both of these are very well located by an object proposal but are usually of no interest. To prevent such objects to be considered as valid tracks, we take the mean optical flow magnitude in all the boxes of the track and we discard it if under a threshold $s$.


\paragraph{Track Ranking}
It is important to compute a score for temporal proposals, in order to account for the likelihood of objects in such proposal. To this end, we propose to consider two factors: the object proposal score used to generate the bounding boxes at each frame and the values given by the IoUs between frames of the tracks. For the former we define $E_t$ as the mean of the scores given by EdgeBoxes, for the latter we define $I_t$ as the mean of all the IoUs of the frames in the track. Using these two figures we define a track score as:
\begin{equation}\label{eq:track-score}
S_t = \lambda E_t~+~(1 - \lambda)I_t
\end{equation}
where $\lambda \in [0, 1]$ is a weighting factor used to balance the contributions of the two scores.


\section{Method Evaluation} \label{entropymethod}
Object proposals are usually evaluated measuring how well objects are covered by the generated boxes. These kind of evaluation does not take into account unannotated objects, and therefore provide a benchmark not reflecting the real capabilities of the proposal method.

The method presented in this paper is a general framework for discovering salient spatio-temporal tracks in videos, which is built upon a generic bounding box oracle. To evaluate it, we introduce a novel method to establish the effectiveness of a generic video proposal, which is also dataset-independent since it does not rely on  annotations. We evaluate whether a proposal effectively represents an instance of some object, since the goal of an object proposal is to locate good candidates and not to produce the candidate of a given class (i.e. the one of the ground truth).
To this end we propose an entropy based evaluation which indicates how the proposal is likely to be recognized as an object. Given a classifier capable of providing for an image a probability distribution $X = \{x_1, \ldots, x_N\}$ over $N$ classes, we compute the Shannon entropy $H$ for the probability vector $X$, 	$H(X) = -\sum_{i=1}^{N} x_i\text{log}(x_i)$.

The rationale behind this choice is that, given a good classifier, for a known object the output probability distribution will be high for the relative class and near zero for the others, thus producing a small entropy. On the contrary, for inputs that the classifier is unsure of, e.g. background patches, the output probability will be distributed non-uniformly among all the possible classes, resulting in a higher entropy.
Therefore, if the classifier is able to cover effectively a sufficiently large number of classes, then the entropy can be interpreted as a measure of \textit{objectness} for the given proposal.


{\setlength\tabcolsep{2pt}
\begin{figure*}[ht]
\begin{tabular}{ccccccc}

\begin{turn}{90}~~~~\scriptsize{Ours}\end{turn}
& \includegraphics[width=.155\textwidth]{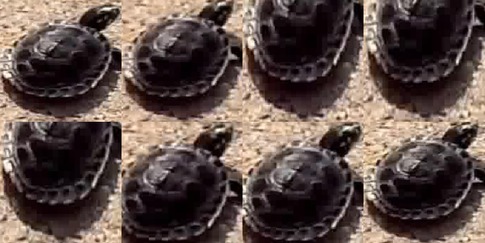}
& \includegraphics[width=.155\textwidth]{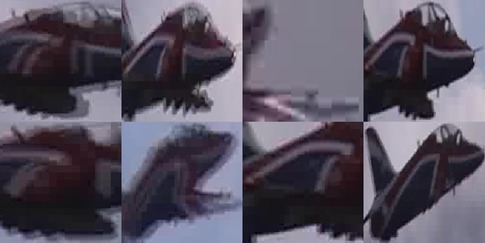}
& \includegraphics[width=.155\textwidth]{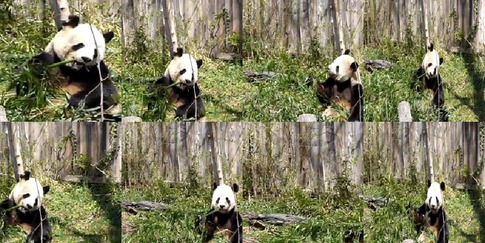}
& \includegraphics[width=.155\textwidth]{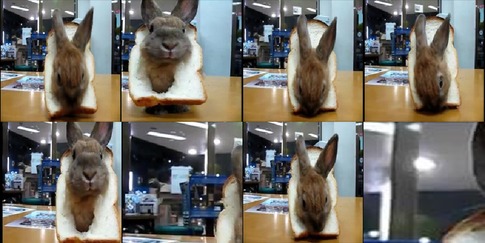}
& \includegraphics[width=.155\textwidth]{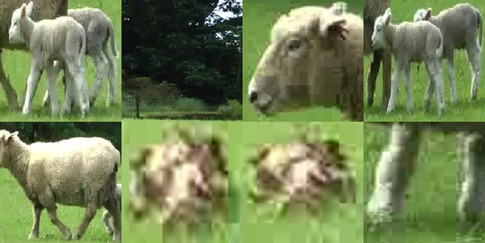}
& \includegraphics[width=.155\textwidth]{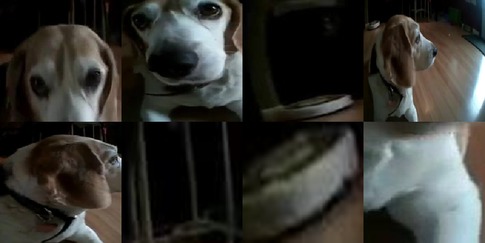} \\

\begin{turn}{90}\tiny{~~}\scriptsize{EdgeBoxes}\end{turn}
&\includegraphics[width=.155\textwidth]{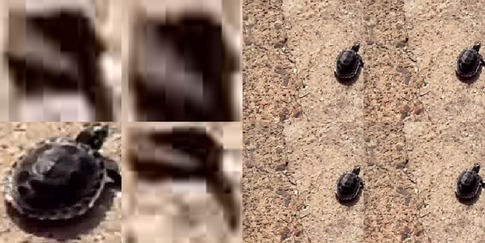}
& \includegraphics[width=.155\textwidth]{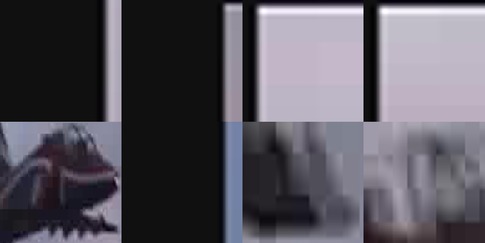}
& \includegraphics[width=.155\textwidth]{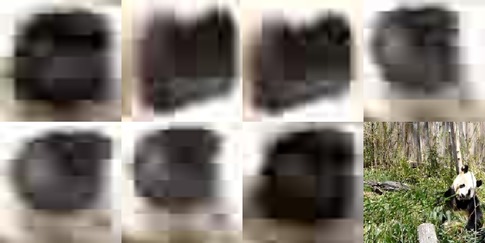}
& \includegraphics[width=.155\textwidth]{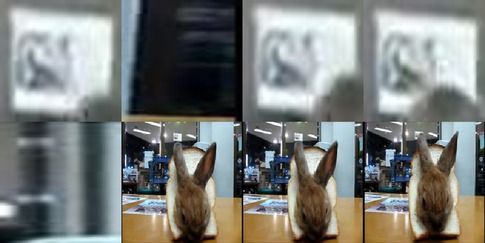}
& \includegraphics[width=.155\textwidth]{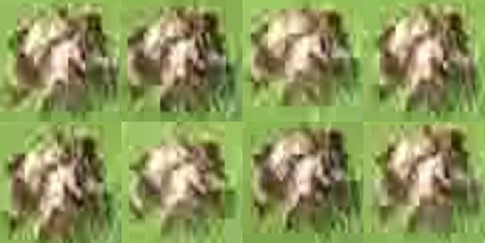}
& \includegraphics[width=.155\textwidth]{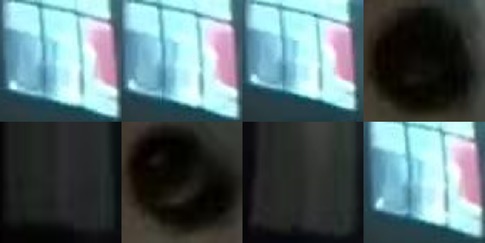} \\

\end{tabular}
\caption{Keyframes of the top 10 tracks in the VID dataset, compared with the top 10 EdgeBoxes proposals. Our method has less redundancy and frames objects more clearly.}
\label{fig:qualitative}
\end{figure*}}

\section{Experiments} \label{experiments}
We experiment on the YouTube-Objects (YTO) \cite{prest2012learning} and on the ILSVRC2015-VID (VID) datasets \cite{russakovsky2015imagenet}, which both provide a per-frame annotation of the objects. YTO is composed by 10 classes and most videos contain a single object per video. VID instead is a more challenging dataset with 30 classes with multiple objects per video.

Here we evaluate our method using the entropy measure introduced in Section \ref{entropymethod}. In all experiments we use EdgeBoxes \cite{zitnick2014edge} as object proposal to generate bounding boxes and as baseline. For the entropy-based proposal scoring we chose the VGG-16 \cite{simonyan2014very} network, trained on the ImageNet \cite{russakovsky2015imagenet} dataset as image classifier, yielding a 1000-dimensional output probability vector. 


For each video we classify 25 proposals and compute the entropy score. For our method we select the best 25 tracks of each video, according to Eq.\ref{eq:track-score}, and for each of them we classify, as representative, the box with the best EdgeBoxes score. We compare the entropy scores against the best 25 boxes given by EdgeBoxes for the whole video.
As a lower-bound reference value we run the classifier on the dataset ground truth. This value is what can be expected to be obtained when proposing only meaningful objects.

Results for YTO are shown in detail in Table \ref{entropy}; it can be seen that our method yields a much lower entropy than EdgeBoxes, also it is close to the ground truth reference. The same trend can be observed on the VID dataset where we measured an average Entropy of 4.73, 3.96 and 3.71 for EdgeBoxes, Our method and the ground truth respectively.

\begin{table}
\centering
\resizebox{\columnwidth}{!}{ 
\begin{tabular}{l|cccccccccc|c}
Method & 
 \includegraphics[width=0.06\columnwidth]{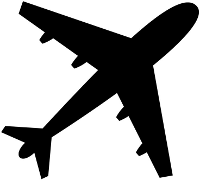}  &
 \includegraphics[width=0.05\columnwidth]{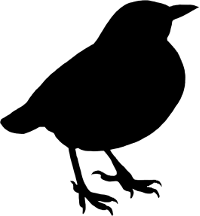}      &
 \includegraphics[width=0.06\columnwidth]{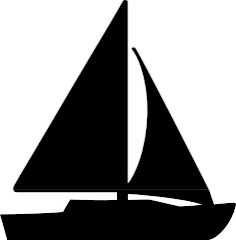}      &
 \includegraphics[width=0.07\columnwidth]{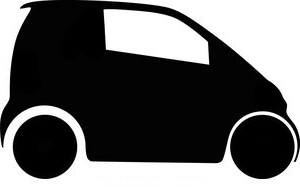}       &
 \includegraphics[width=0.08\columnwidth]{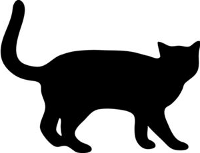}       &
 \includegraphics[width=0.08\columnwidth]{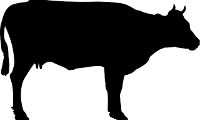}       &
 \includegraphics[width=0.08\columnwidth]{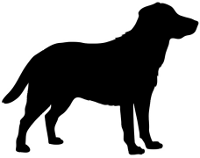}       &
 \includegraphics[width=0.08\columnwidth]{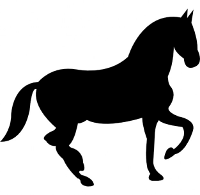}     &
 \includegraphics[width=0.10\columnwidth]{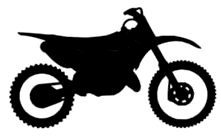} &
\includegraphics[width=0.08\columnwidth]{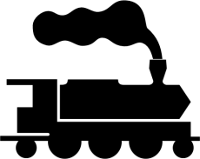}     &
Mean\\
\hline

\hline
EB \cite{zitnick2014edge} & 5.02 & 5.19 & 5.48 & 4.52 & 5.92 & 6.27 & 6.16 & 6.54 & 5.68 & 5.19 & 5.60 \Tstrut \\
\hline
Ours & 3.58 & 3.25 & 3.10 & 2.45 & 4.02 & 3.00 & 3.58 & 3.25 & 3.10 & 2.45 & 3.18 \Tstrut \\
\hline
\hline
GT & 0.58 & 1.33 & 1.03 & 1.83 & 2.31 & 2.41 & 2.28 & 2.58 & 2.57 & 2.41 & 1.93
\Tstrut \\
\hline

\hline
\end{tabular}}
\caption{Entropy comparison (lower is better) between the proposals provided by EdgeBoxes (EB) and our method (Ours) and Ground Truth boxes (GT).}
\label{entropy}
\end{table}
\paragraph{High precision proposals}
As a further evaluation, we treated our proposal as an object detector measuring the mean Average Precision (mAP) for the YTO dataset. This aims at measuring the precision of a proposal method. Since the class set of YTO is a subset of the one of Pascal VOC \cite{everingham2010pascal}, for this evaluation we used Fast-RCNN \cite{girshick2015fast}, restricted to the ten common classes.
 
Table \ref{apcomp} shows a comparison between our proposed tracks and EdgeBoxes. In order to make the comparison fair, we evaluated the best 25 boxes proposed by both methods for each video, similarly to the entropy evaluation in Section \ref{experiments}. The mAP of our tracks is 8.5 times higher than EdgeBoxes, proving that our proposal is much more precise. 

\paragraph{Qualitative results}
We report some qualitative results, showing a comparison of content extracted by our proposal with respect to EdgeBoxes. We compare the best boxes and tracks in a given video. In Fig.~\ref{fig:qualitative} it can be seen how our proposals are more diverse and frame an object correctly with respect to the top proposal chosen from EdgeBoxes.
 
In Fig.~\ref{fig:qualitative-entropy} we show an example of high and low entropy proposals. For our method and EdgeBoxes we report the 10 boxes with the lowest and highest entropies among the first best 25 proposals. As reference we also report high and low entropy boxes from the ground truth. It can be seen that our method is more focused on objects even in its highest entropy proposals.

\begin{table}
\centering
\resizebox{\columnwidth}{!}{ 
\begin{tabular}{l|cccccccccc|c}
Method & 
\includegraphics[width=0.06\columnwidth]{images/icons/airplane.png}  &
\includegraphics[width=0.05\columnwidth]{images/icons/bird.png}      &
\includegraphics[width=0.06\columnwidth]{images/icons/boat.png}      &
\includegraphics[width=0.07\columnwidth]{images/icons/car.png}       &
\includegraphics[width=0.08\columnwidth]{images/icons/cat.png}       &
\includegraphics[width=0.08\columnwidth]{images/icons/cow.png}       &
\includegraphics[width=0.08\columnwidth]{images/icons/dog.png}       &
\includegraphics[width=0.08\columnwidth]{images/icons/horse.png}     &
\includegraphics[width=0.10\columnwidth]{images/icons/motorbike.png} &
\includegraphics[width=0.08\columnwidth]{images/icons/train.png}     &
Mean\\
\hline

\hline
EB \cite{zitnick2014edge} &
0.94 & 0.40 & 0.49 & ~~1.80 & 10.96 & 0.57 & 0.56 & 0.61 & 0.26 & ~~2.95 & 0.98 \Tstrut \\
\hline
Ours &
9.15 & 7.16 & 5.98 &  14.94 & 10.95 & 8.43 & 6.10 & 2.26 & 3.42 &  14.91 & 8.33 \Tstrut \\
\hline

\hline
\end{tabular}}
\caption{AP comparison (higher is better) for object detection between EdgeBoxes (EB) and Our method.}
\label{apcomp}
\end{table}

\newcommand{\entfigsize}{.09\columnwidth}
{\setlength\tabcolsep{2pt}
\begin{figure}
\centering
\resizebox{\columnwidth}{!}{%
\begin{tabular}{c *{9}{c}}

 & \multicolumn{9}{c}{\scriptsize{Low Entropy}} \\

\begin{turn}{90}~~\scriptsize{EB}\end{turn}
& \includegraphics[width=\entfigsize, height=\entfigsize]{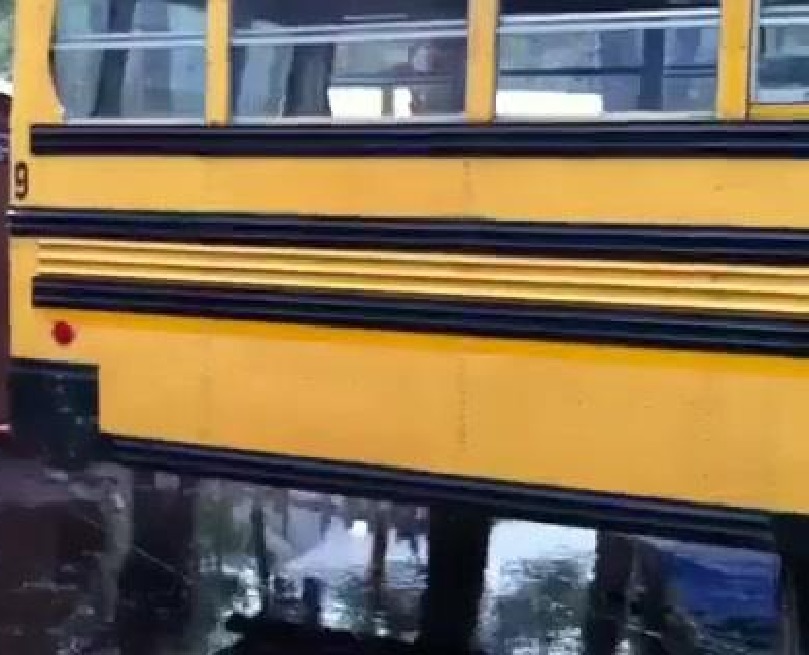}
& \includegraphics[width=\entfigsize, height=\entfigsize]{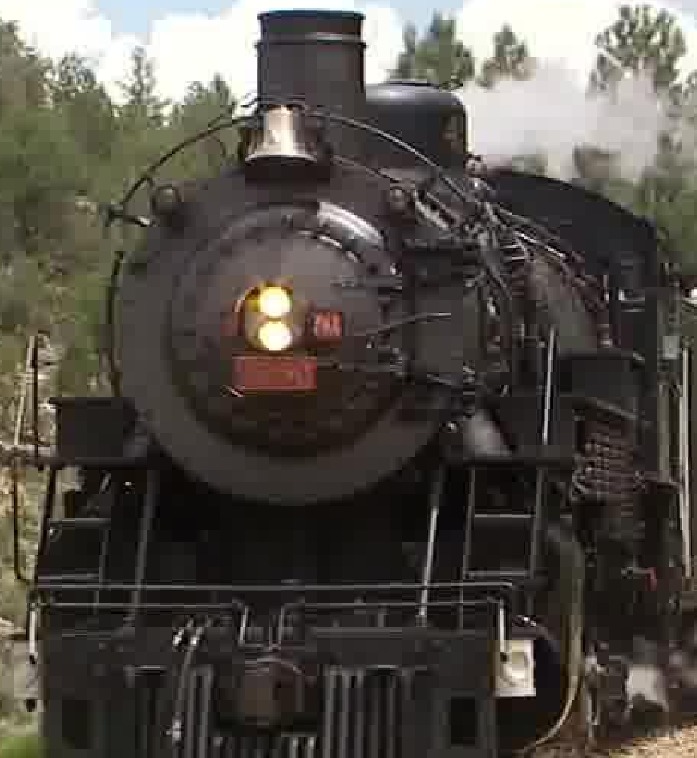}
& \includegraphics[width=\entfigsize, height=\entfigsize]{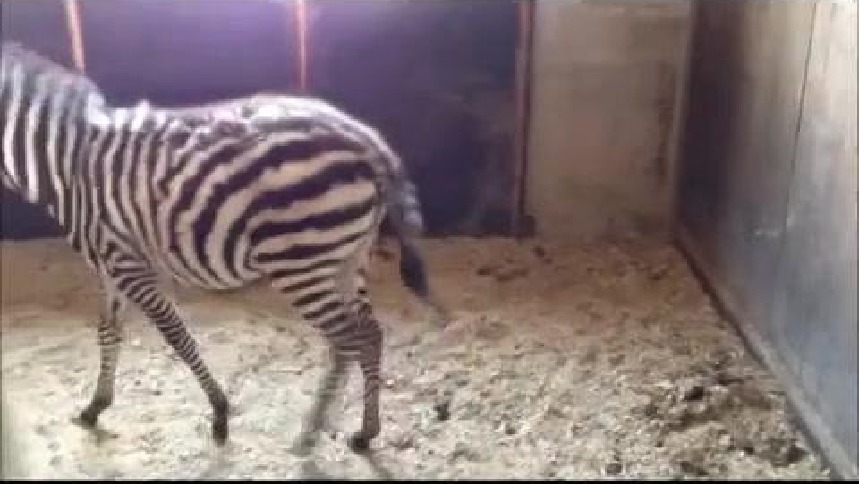}
& \includegraphics[width=\entfigsize, height=\entfigsize]{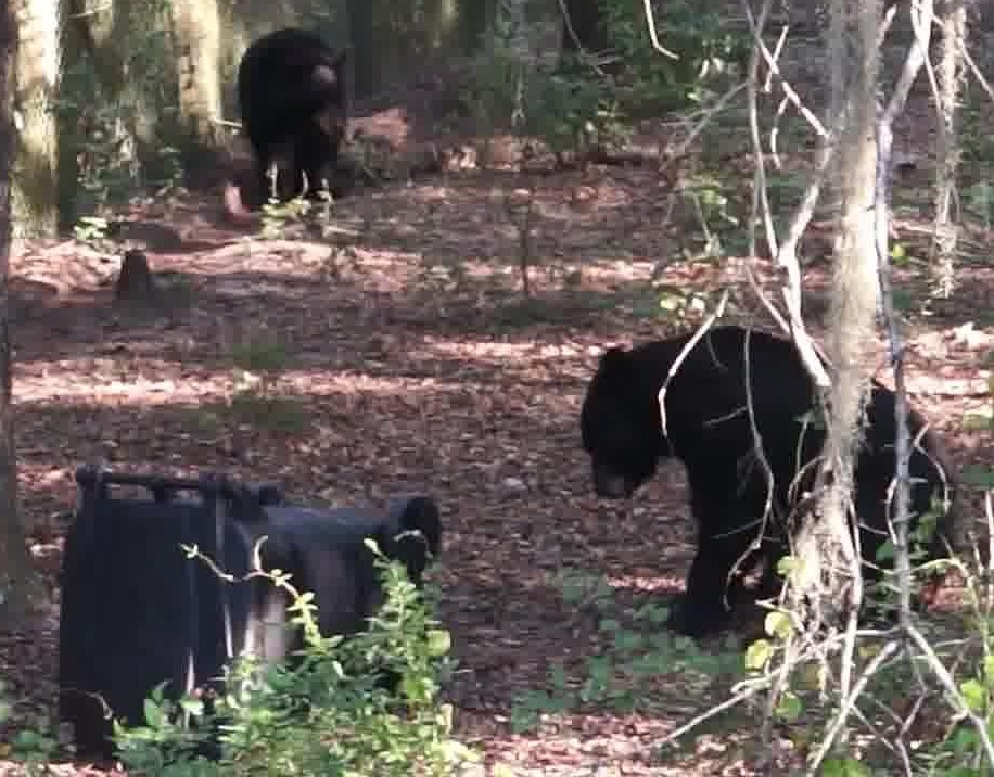}
& \includegraphics[width=\entfigsize, height=\entfigsize]{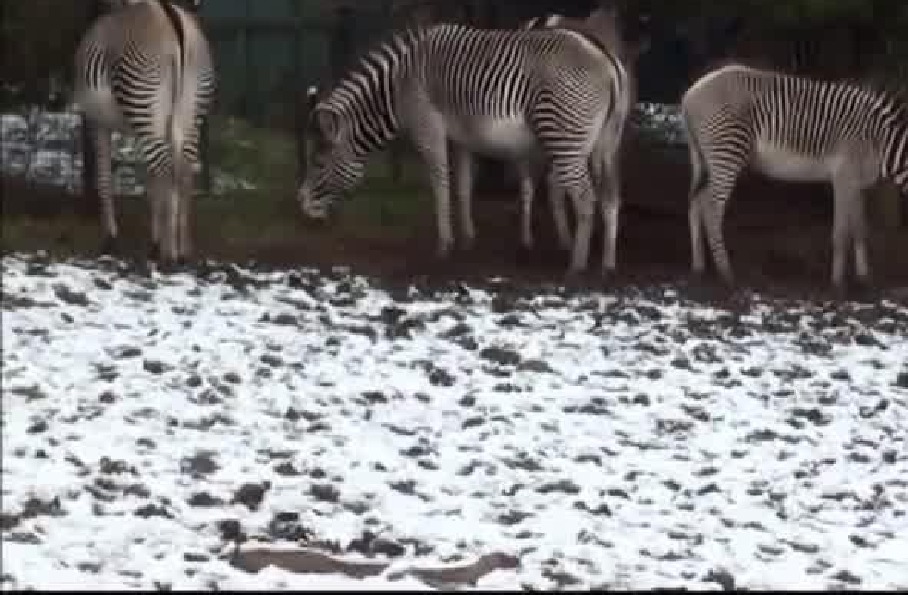}
& \includegraphics[width=\entfigsize, height=\entfigsize]{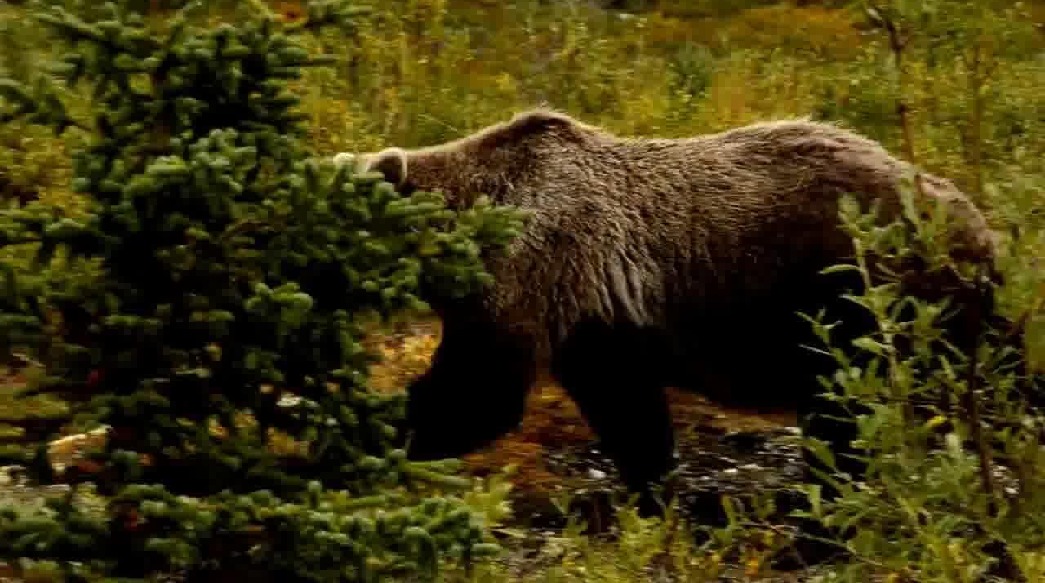}
& \includegraphics[width=\entfigsize, height=\entfigsize]{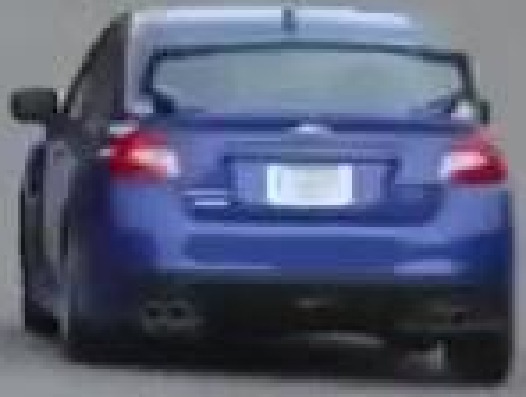}
& \includegraphics[width=\entfigsize, height=\entfigsize]{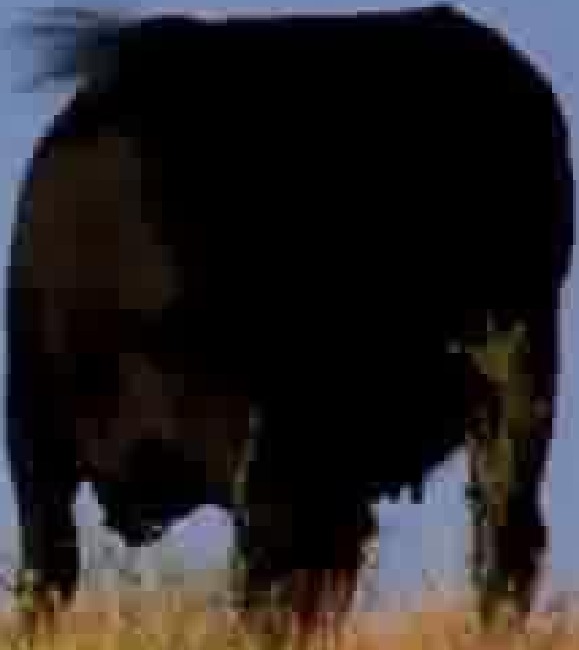}
& \includegraphics[width=\entfigsize, height=\entfigsize]{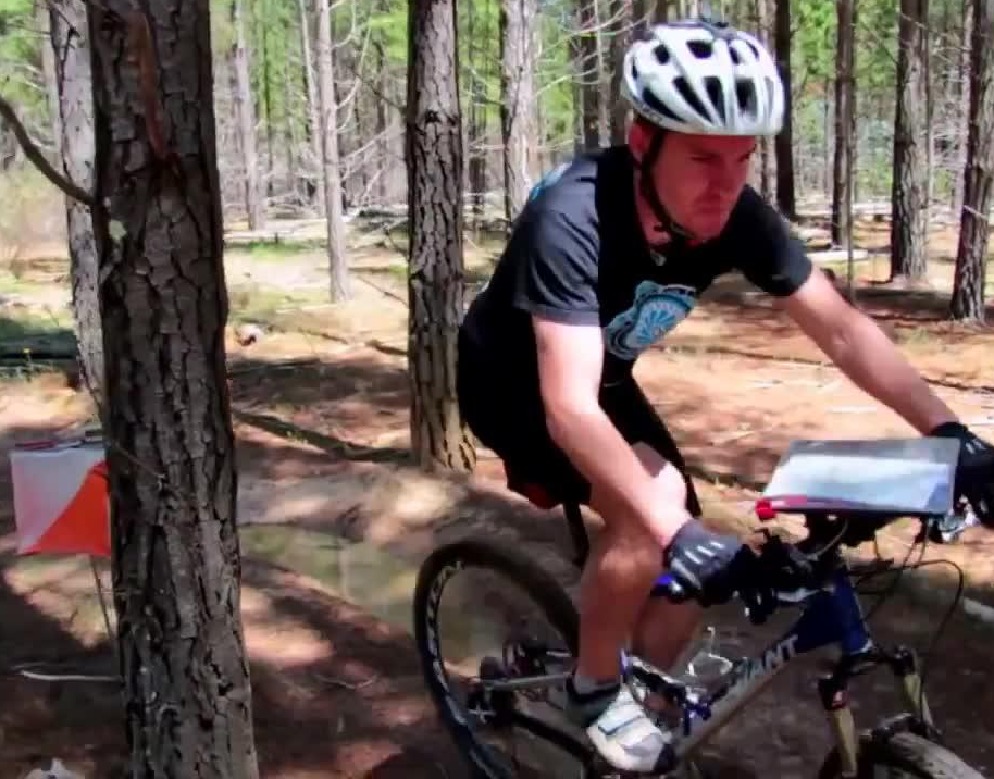}
\\


\begin{turn}{90}~\scriptsize{Ours}~\end{turn}
& \includegraphics[width=\entfigsize, height=\entfigsize]{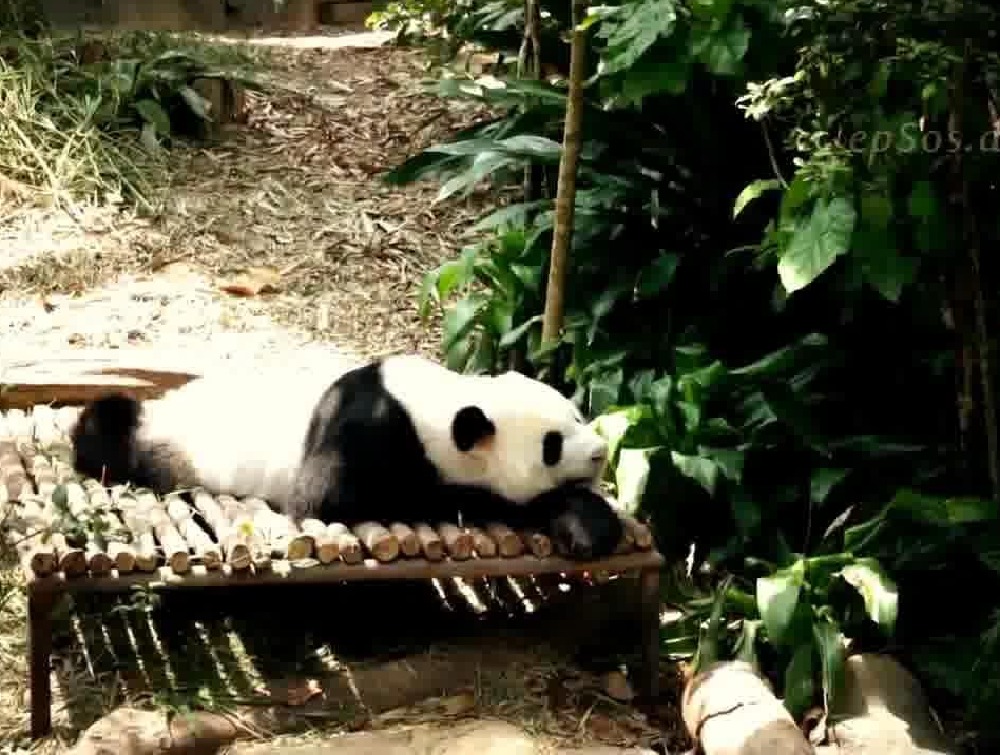}
& \includegraphics[width=\entfigsize, height=\entfigsize]{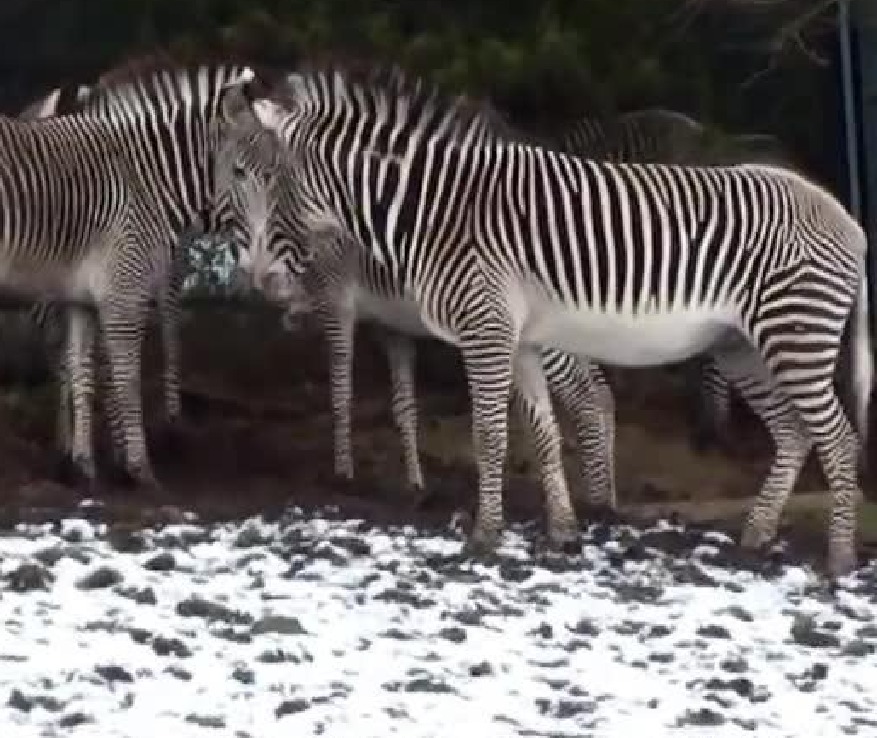}
& \includegraphics[width=\entfigsize, height=\entfigsize]{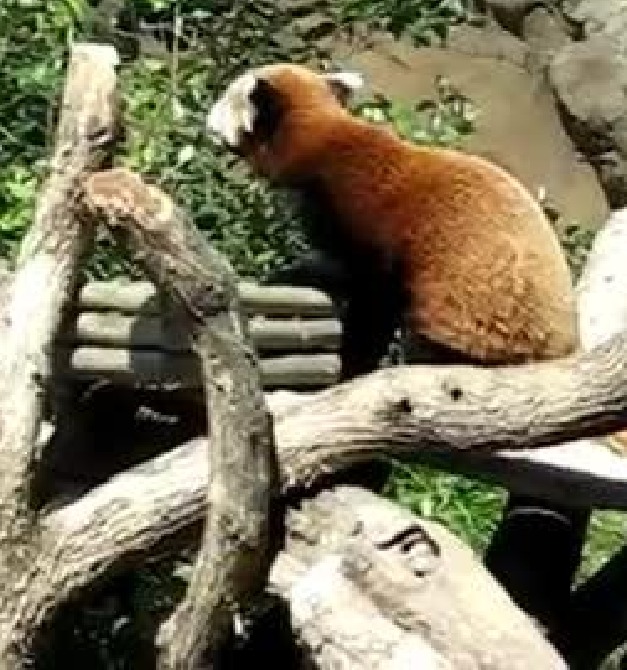}
& \includegraphics[width=\entfigsize, height=\entfigsize]{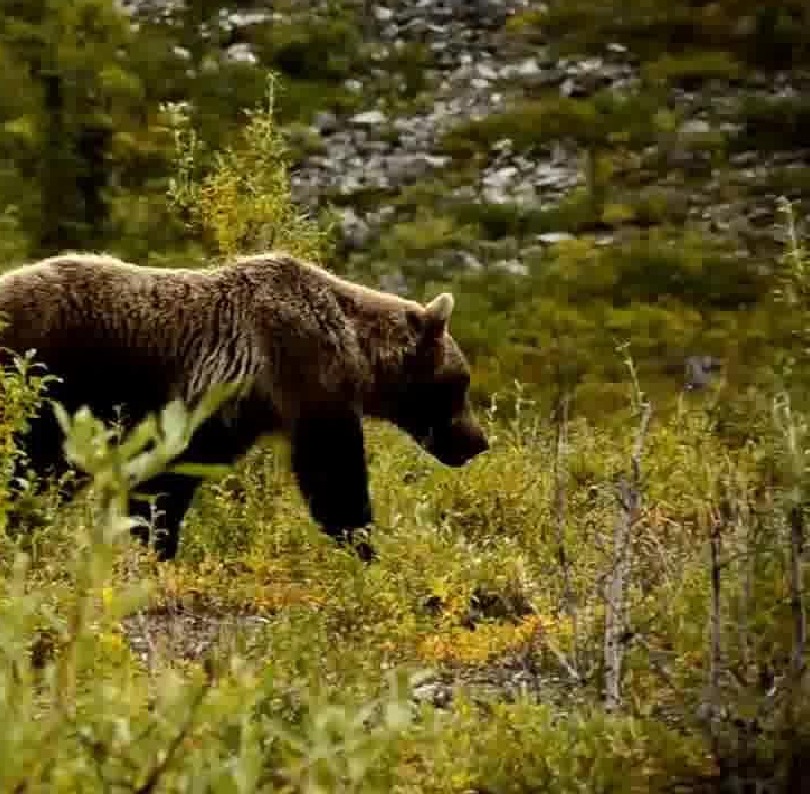}
& \includegraphics[width=\entfigsize, height=\entfigsize]{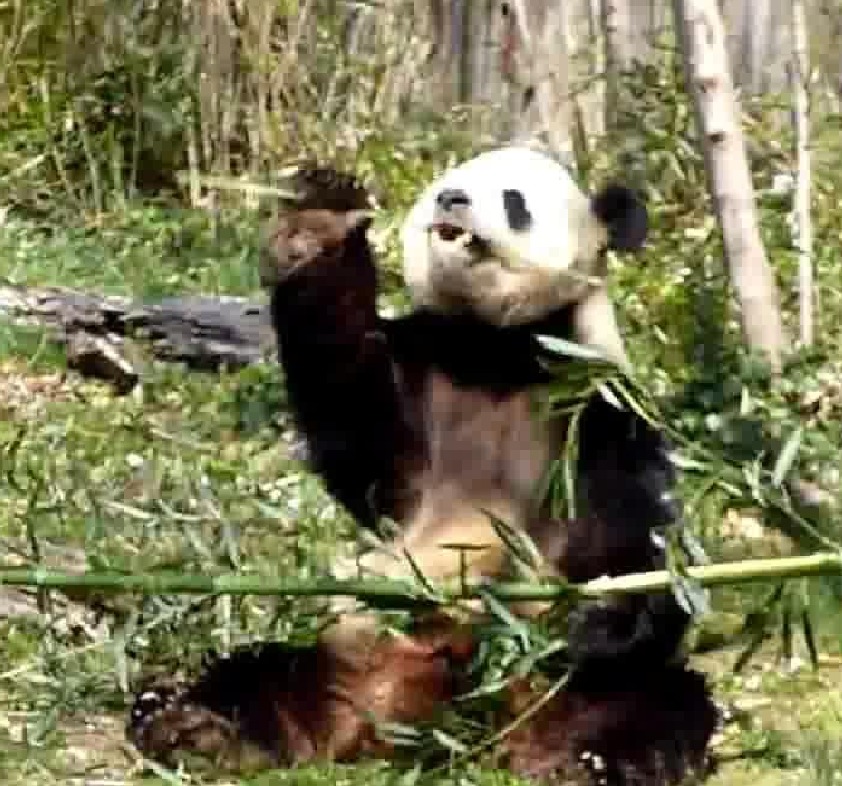}
& \includegraphics[width=\entfigsize, height=\entfigsize]{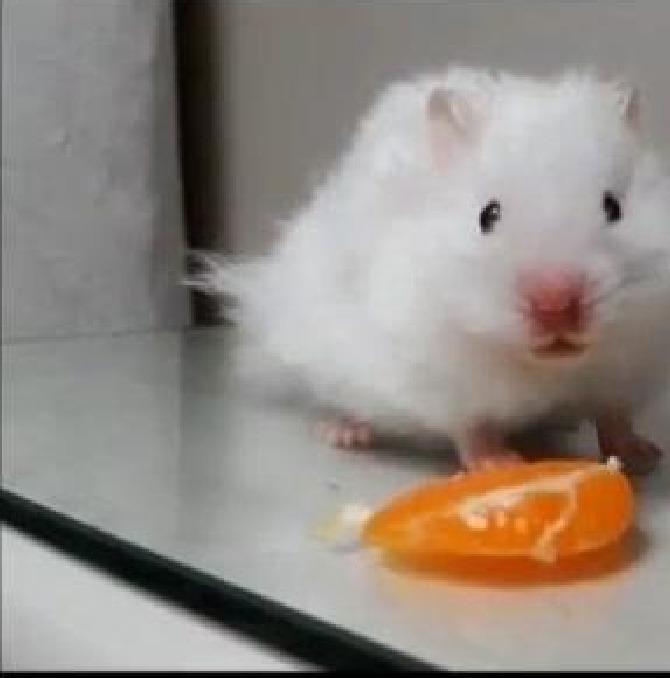}
& \includegraphics[width=\entfigsize, height=\entfigsize]{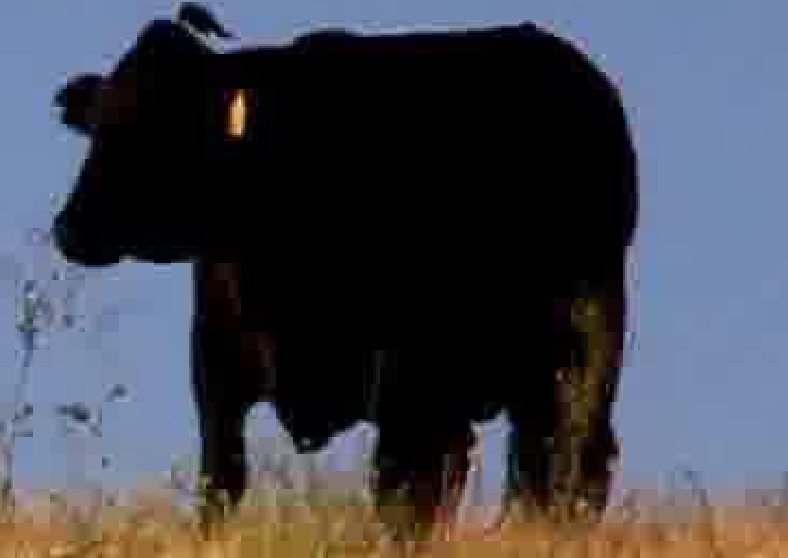}
& \includegraphics[width=\entfigsize, height=\entfigsize]{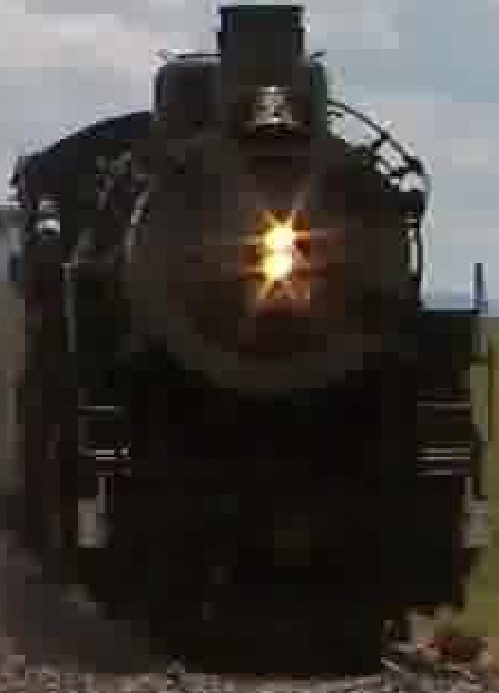}
& \includegraphics[width=\entfigsize, height=\entfigsize]{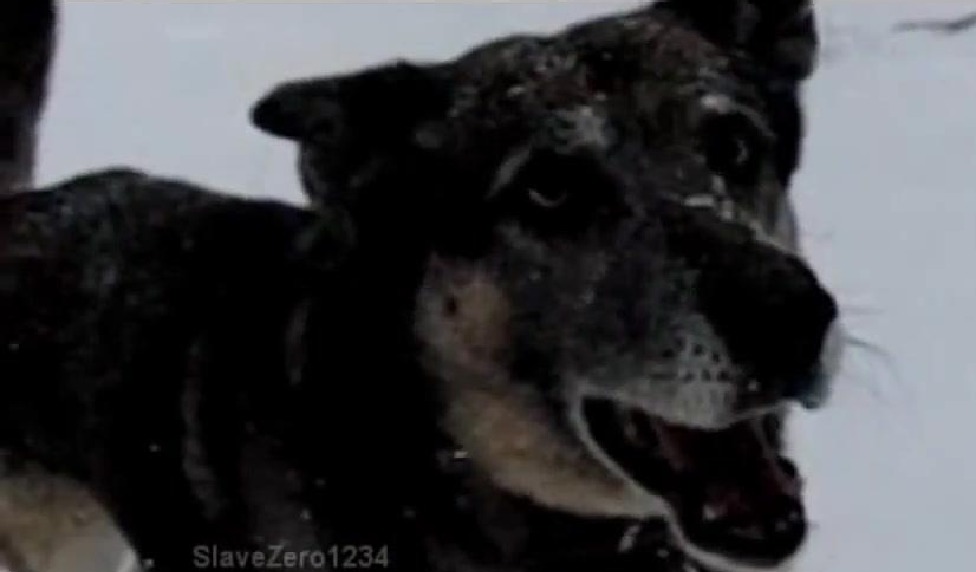}
\\


\begin{turn}{90}~~\scriptsize{GT}\end{turn}
& \includegraphics[width=\entfigsize, height=\entfigsize]{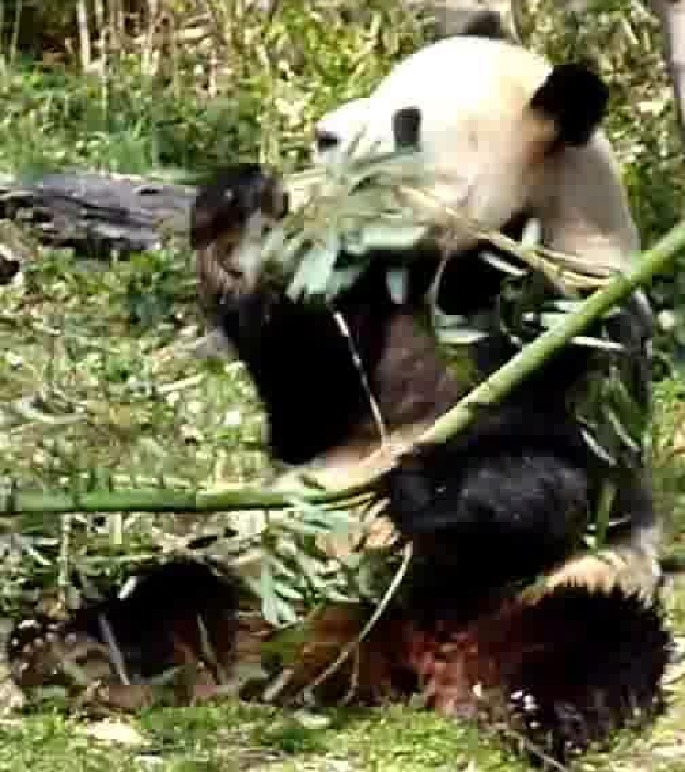}
& \includegraphics[width=\entfigsize, height=\entfigsize]{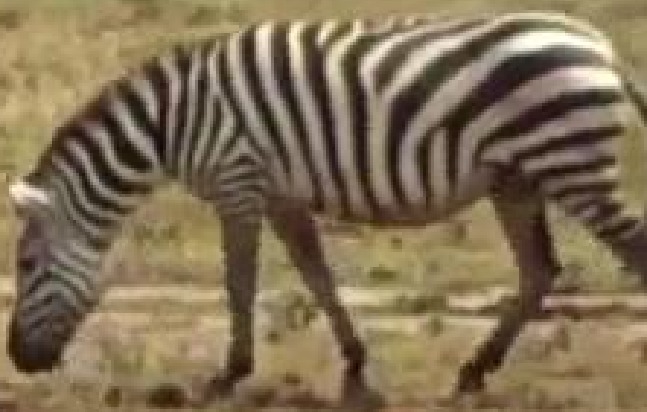}
& \includegraphics[width=\entfigsize, height=\entfigsize]{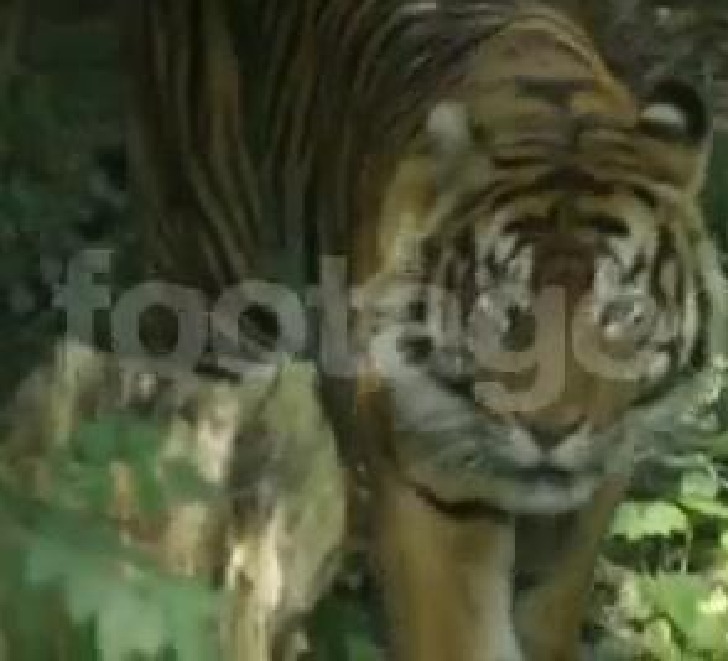}
& \includegraphics[width=\entfigsize, height=\entfigsize]{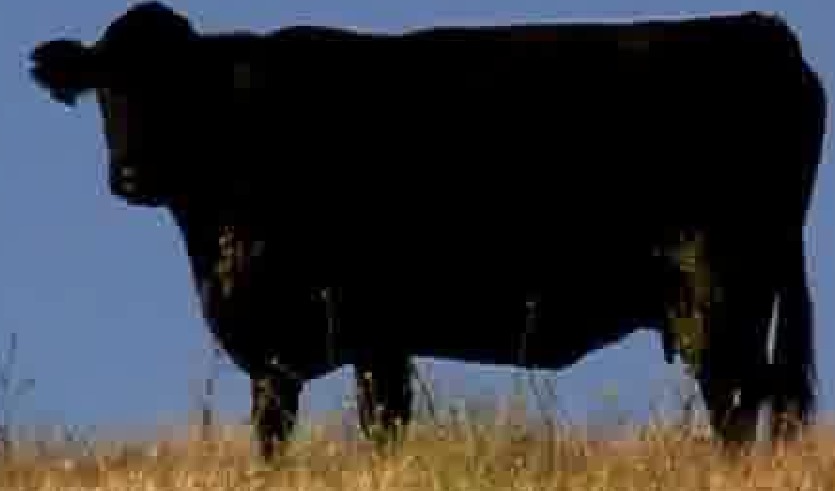}
& \includegraphics[width=\entfigsize, height=\entfigsize]{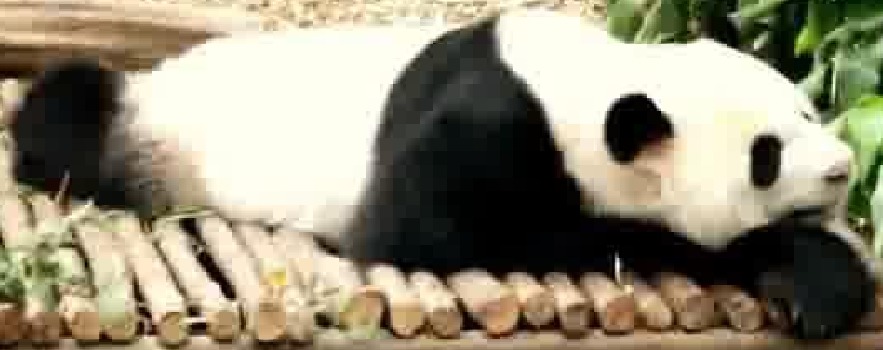}
& \includegraphics[width=\entfigsize, height=\entfigsize]{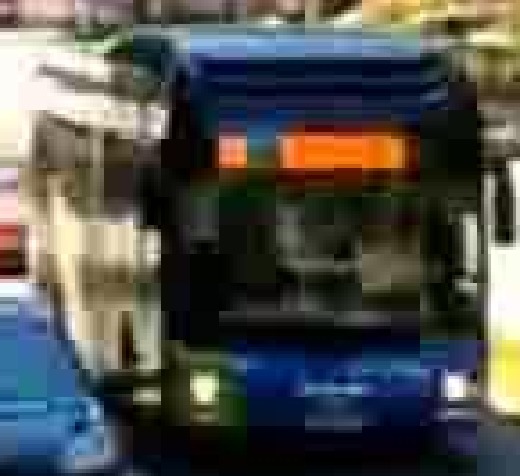}
& \includegraphics[width=\entfigsize, height=\entfigsize]{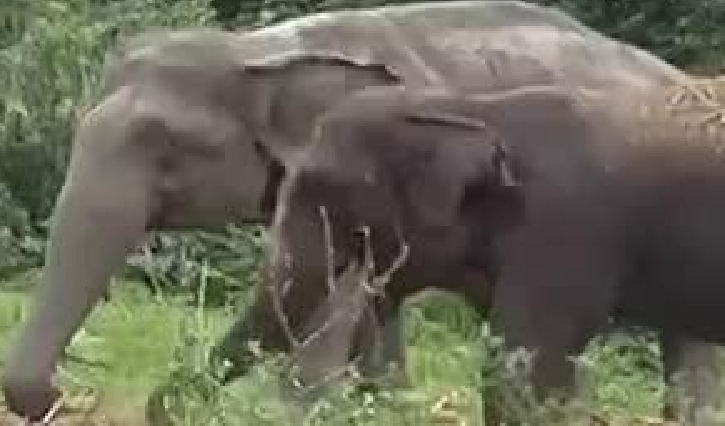}
& \includegraphics[width=\entfigsize, height=\entfigsize]{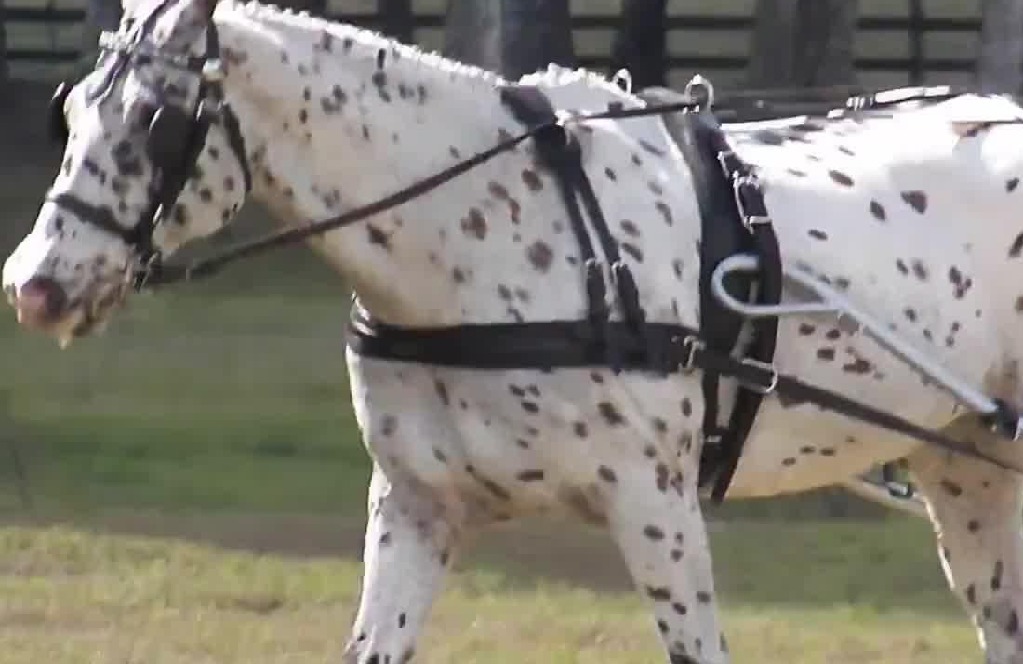}
& \includegraphics[width=\entfigsize, height=\entfigsize]{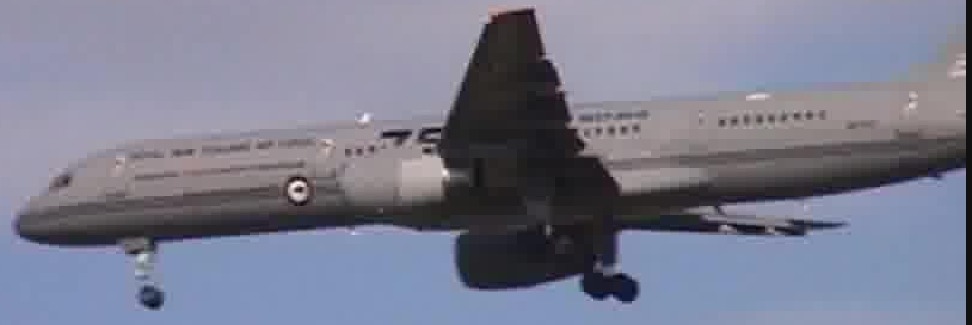}
\\
& \multicolumn{9}{c}{\scriptsize{High Entropy}} \\
\begin{turn}{90}~~\scriptsize{EB}\end{turn} & \includegraphics[width=\entfigsize, height=\entfigsize]{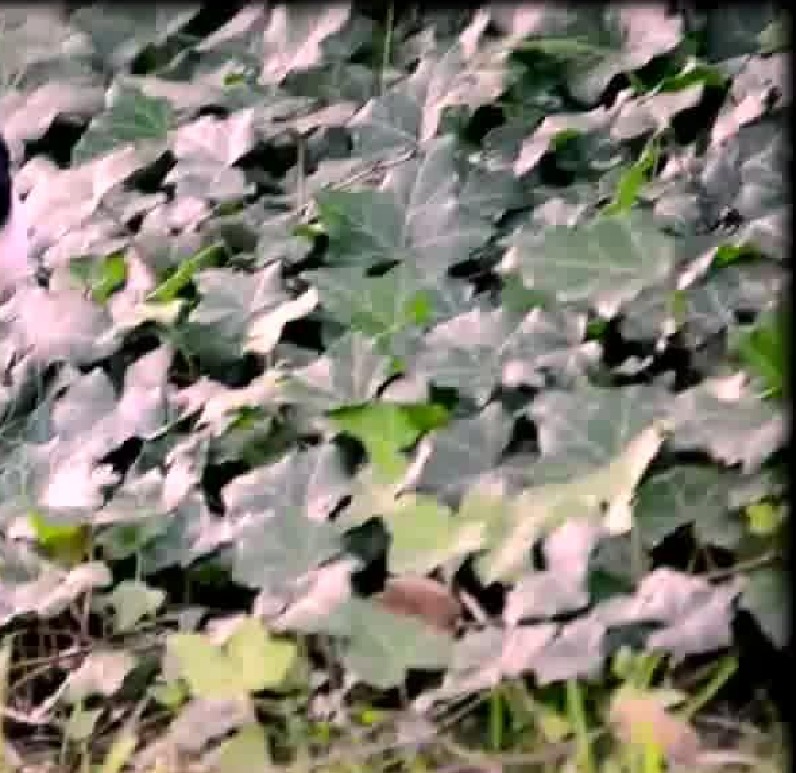}
& \includegraphics[width=\entfigsize, height=\entfigsize]{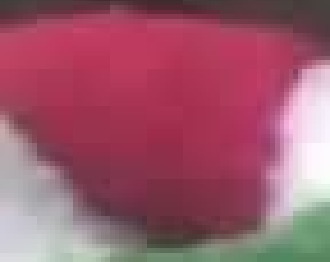}
& \includegraphics[width=\entfigsize, height=\entfigsize]{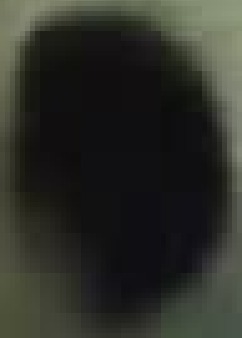}
& \includegraphics[width=\entfigsize, height=\entfigsize]{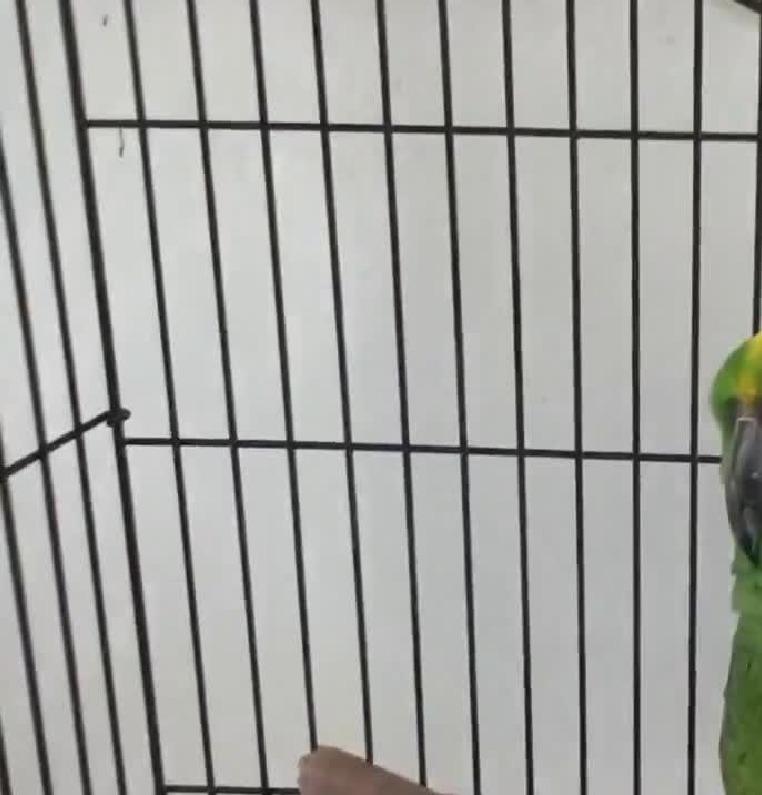}
& \includegraphics[width=\entfigsize, height=\entfigsize]{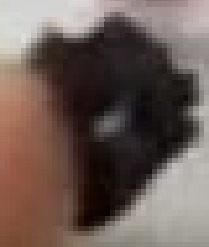}
& \includegraphics[width=\entfigsize, height=\entfigsize]{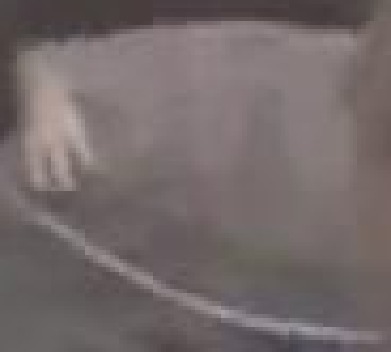}
& \includegraphics[width=\entfigsize, height=\entfigsize]{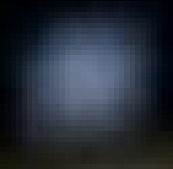}
& \includegraphics[width=\entfigsize, height=\entfigsize]{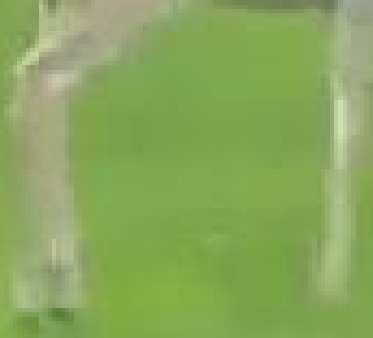}
& \includegraphics[width=\entfigsize, height=\entfigsize]{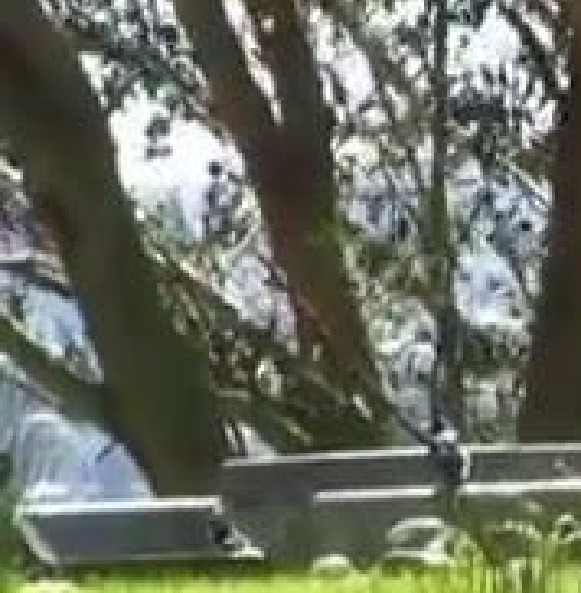}
\\

\begin{turn}{90}~\scriptsize{Ours}\end{turn} & \includegraphics[width=\entfigsize, height=\entfigsize]{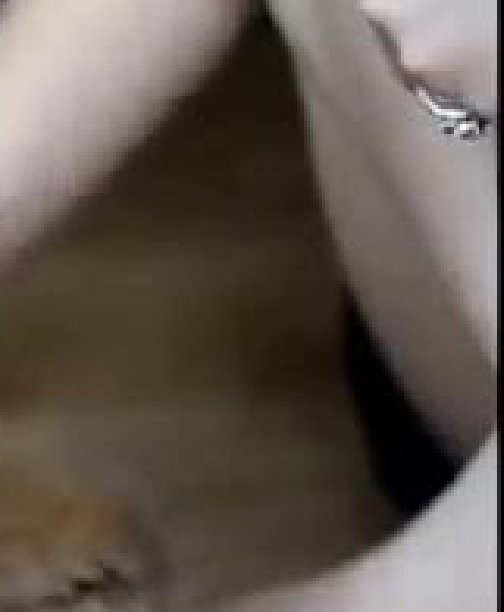}
& \includegraphics[width=\entfigsize, height=\entfigsize]{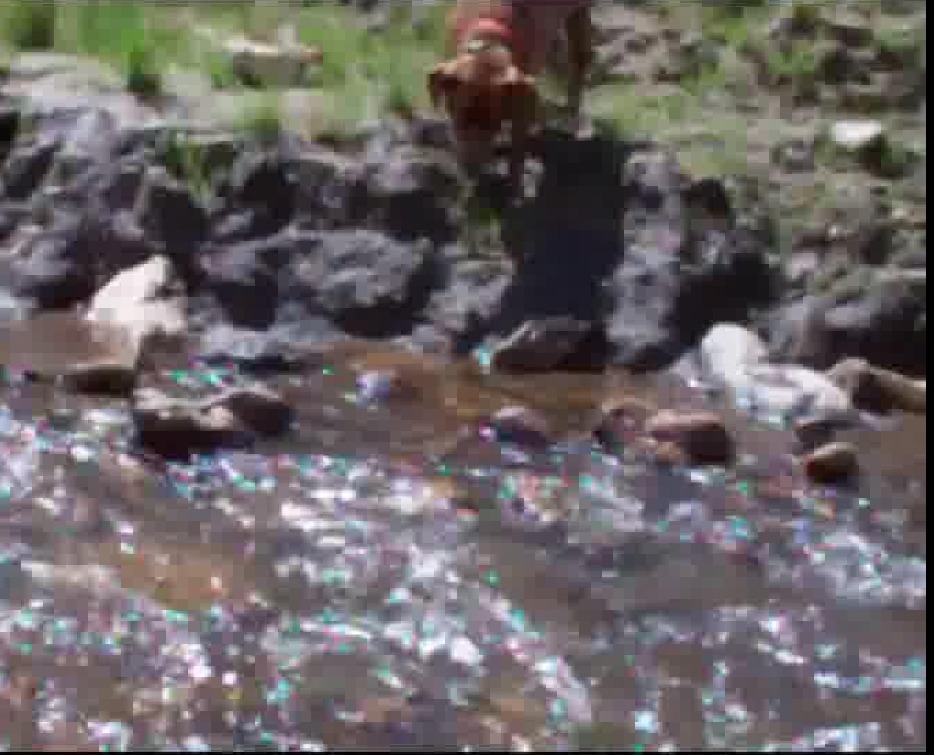}
& \includegraphics[width=\entfigsize, height=\entfigsize]{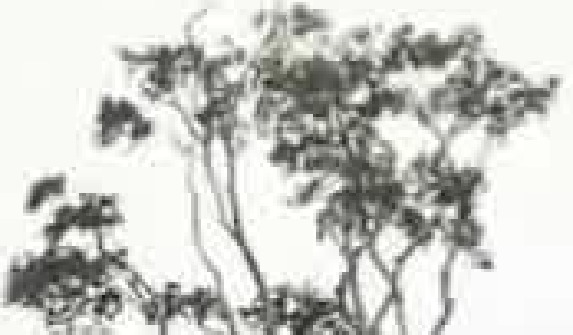}
& \includegraphics[width=\entfigsize, height=\entfigsize]{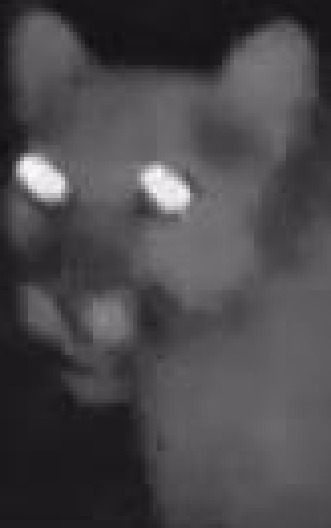}
& \includegraphics[width=\entfigsize, height=\entfigsize]{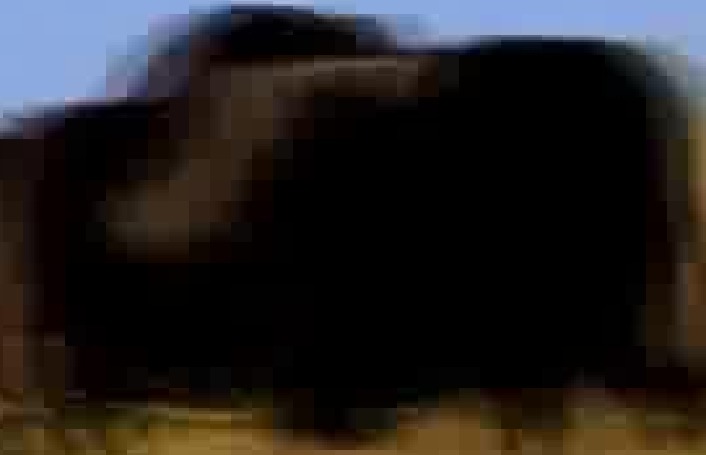}
& \includegraphics[width=\entfigsize, height=\entfigsize]{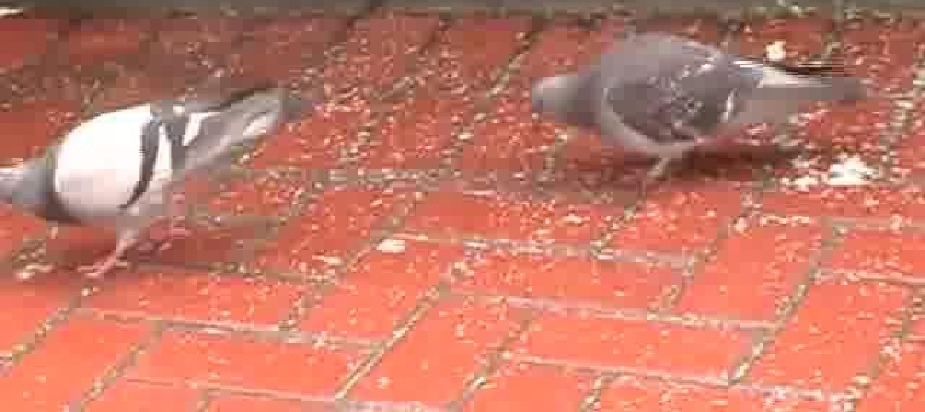}
& \includegraphics[width=\entfigsize, height=\entfigsize]{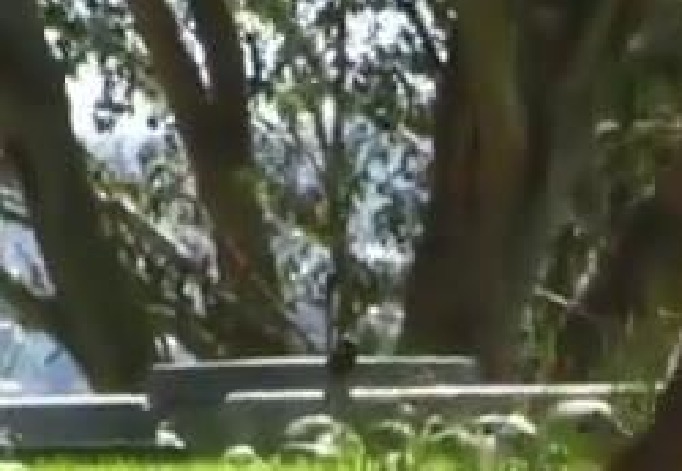}
& \includegraphics[width=\entfigsize, height=\entfigsize]{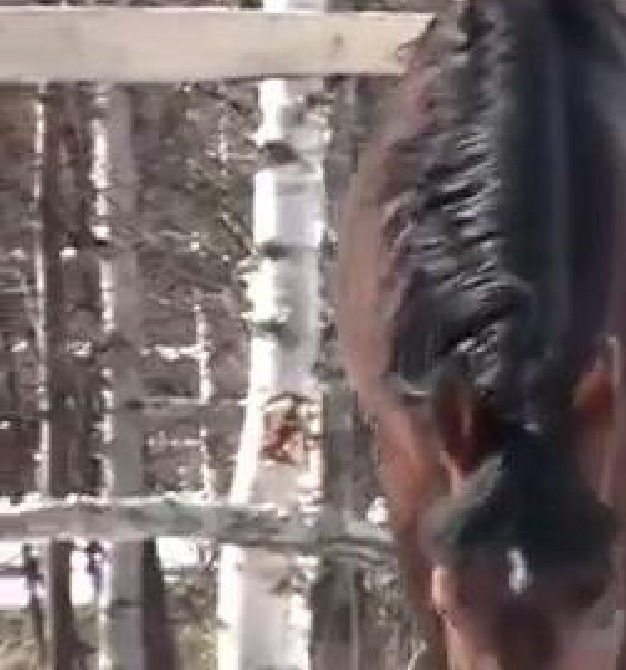}
& \includegraphics[width=\entfigsize, height=\entfigsize]{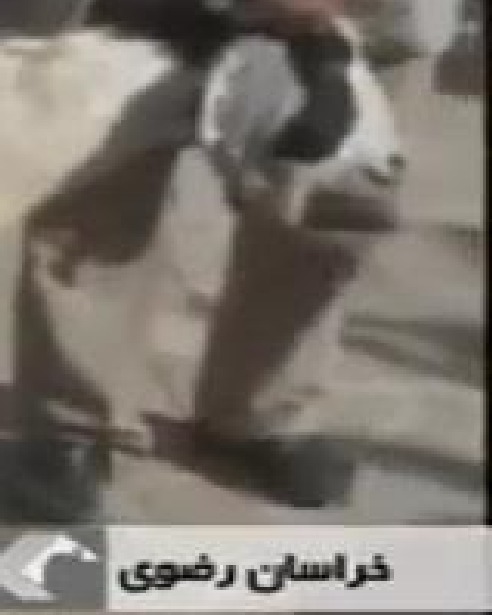}
\\

\begin{turn}{90}~~\scriptsize{GT}\end{turn} & \includegraphics[width=\entfigsize, height=\entfigsize]{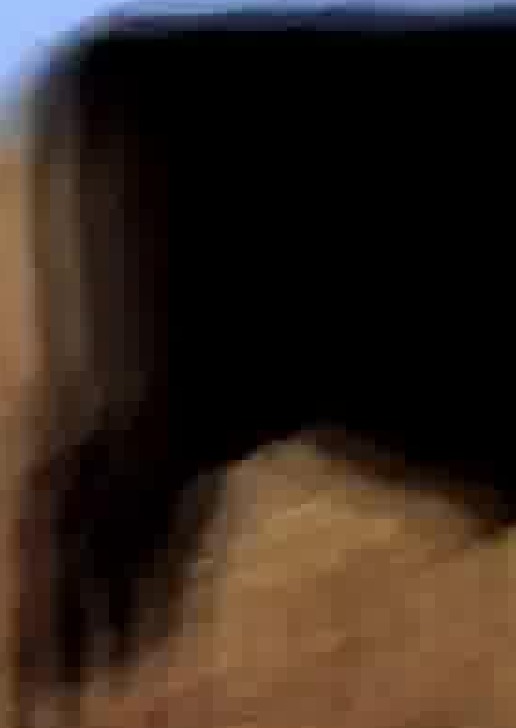}
& \includegraphics[width=\entfigsize, height=\entfigsize]{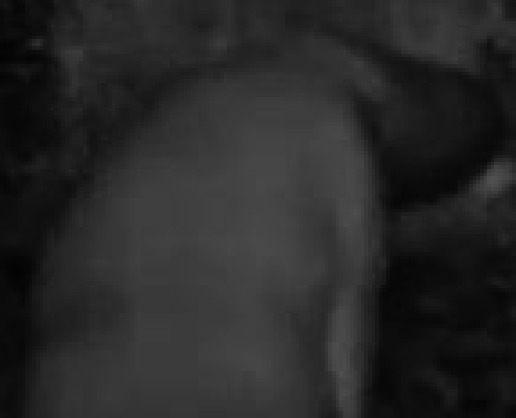}
& \includegraphics[width=\entfigsize, height=\entfigsize]{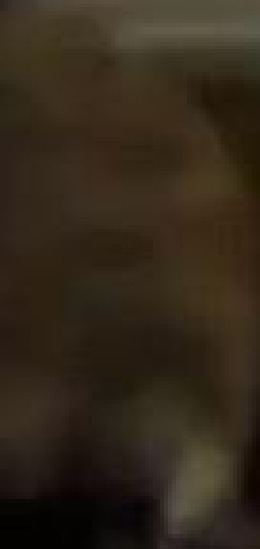}
& \includegraphics[width=\entfigsize, height=\entfigsize]{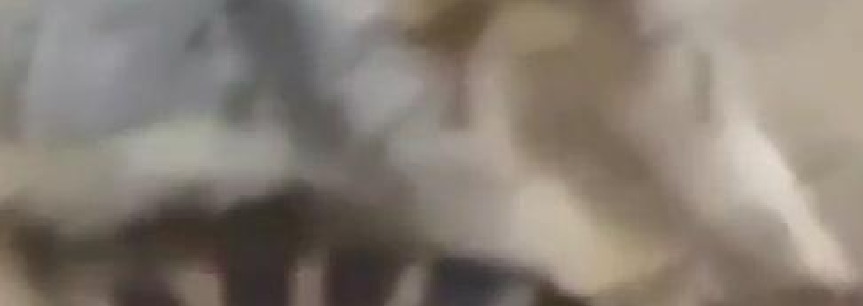}
& \includegraphics[width=\entfigsize, height=\entfigsize]{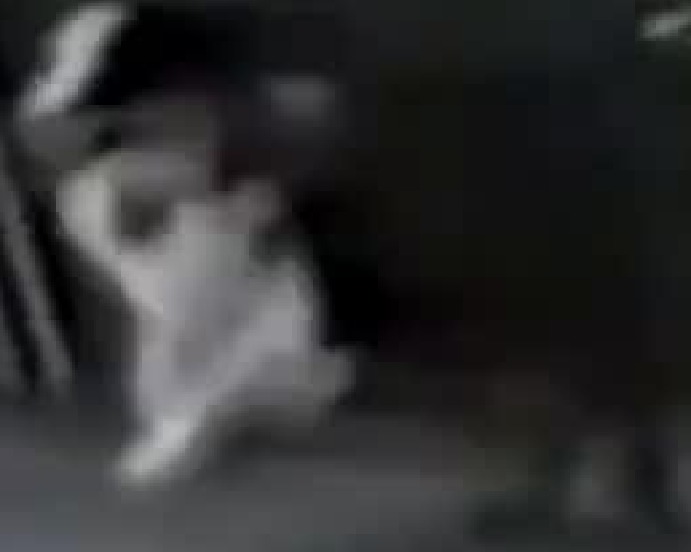}
& \includegraphics[width=\entfigsize, height=\entfigsize]{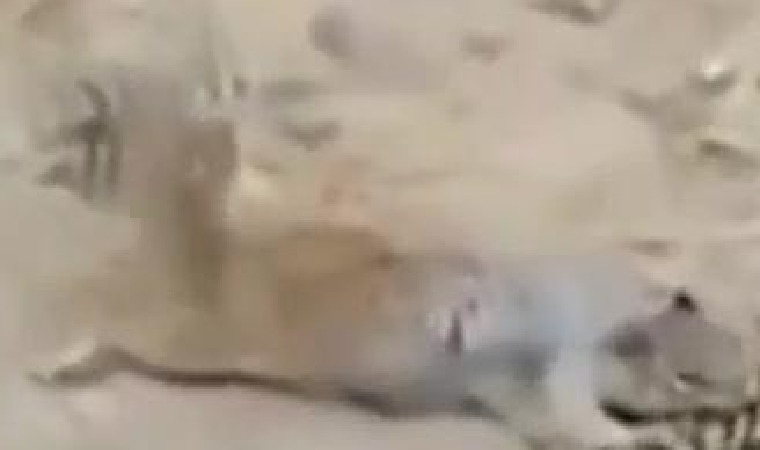}
& \includegraphics[width=\entfigsize, height=\entfigsize]{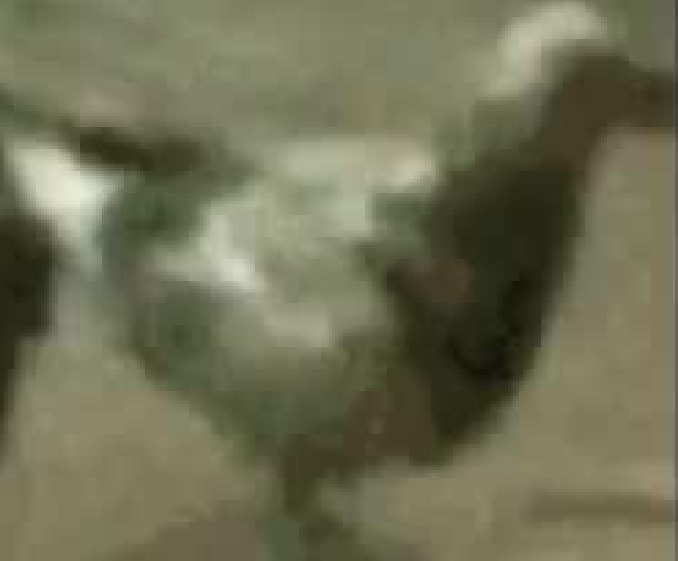}
& \includegraphics[width=\entfigsize, height=\entfigsize]{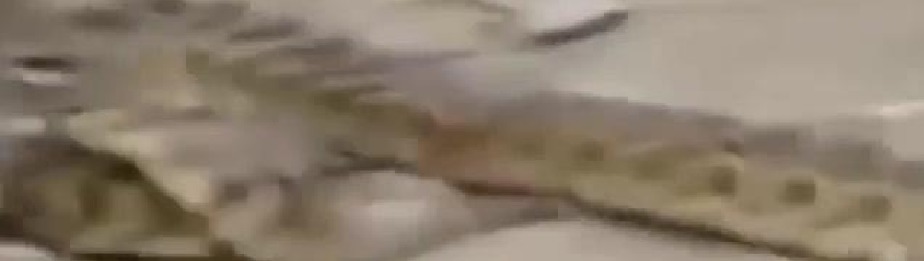}
& \includegraphics[width=\entfigsize, height=\entfigsize]{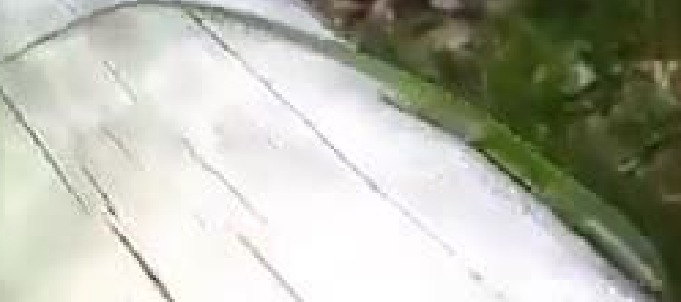}
\end{tabular}
}
\caption{Highest and lowest entropy proposals for our method (Ours), EdgeBoxes (EB) and ground truth boxes (GT).}
\label{fig:qualitative-entropy}
\end{figure}}

\section{Conclusions} \label{conclusions}
We proposed a novel and unsupervised method to extract from videos, tracks containing meaningful objects. Our track proposal can build on any  object bounding box proposal method. The matching process only relies on bounding box geometry and optical flow, resulting in a simple and effective method for high precision video object proposals. We also introduce a dataset independent method to evaluate the effectiveness of an object proposal, not relying on dataset annotations. The proposal has been evaluated on the YouTube Objects and ILSVRC-2015 VID datasets, showing a high precision and providing meaningful object proposals that can be used for any video analysis task without looking at the whole sequence.


{\footnotesize
\bibliographystyle{plain}
\bibliography{egbib}

\begin{thebibliography}{10}

\bibitem{everingham2010pascal}
Mark Everingham, Luc Van~Gool, Christopher~KI Williams, John Winn, and Andrew
  Zisserman.
\newblock The pascal visual object classes (voc) challenge.
\newblock {\em International journal of computer vision}, 88(2):303--338, 2010.

\bibitem{girshick2015fast}
Ross Girshick.
\newblock Fast r-cnn.
\newblock In {\em Proc. of ICCV}, 2015.

\bibitem{hosang2015}
Jan Hosang, Rodrigo Benenson, Piotr Doll{\'a}r, and Bernt Schiele.
\newblock What makes for effective detection proposals?
\newblock {\em IEEE Transactions on Pattern Analysis and Machine Intelligence},
  38(4):814--830, 2016.

\bibitem{kwak2015unsupervised}
Suha Kwak, Minsu Cho, Ivan Laptev, Jean Ponce, and Cordelia Schmid.
\newblock Unsupervised object discovery and tracking in video collections.
\newblock In {\em Proc. of ICCV}, 2015.

\bibitem{oneata2014spatio}
Dan Oneata, J{\'e}r{\^o}me Revaud, Jakob Verbeek, and Cordelia Schmid.
\newblock Spatio-temporal object detection proposals.
\newblock In {\em Proc. of ECCV}. Springer, 2014.

\bibitem{prest2012learning}
Alessandro Prest, Christian Leistner, Javier Civera, Cordelia Schmid, and
  Vittorio Ferrari.
\newblock Learning object class detectors from weakly annotated video.
\newblock In {\em Proc. of CVPR}. IEEE, 2012.

\bibitem{russakovsky2015imagenet}
Olga Russakovsky, Jia Deng, Hao Su, Jonathan Krause, Sanjeev Satheesh, Sean Ma,
  Zhiheng Huang, Andrej Karpathy, Aditya Khosla, Michael Bernstein, et~al.
\newblock Imagenet large scale visual recognition challenge.
\newblock {\em International Journal of Computer Vision}, 115(3):211--252,
  2015.

\bibitem{simonyan2014very}
Karen Simonyan and Andrew Zisserman.
\newblock Very deep convolutional networks for large-scale image recognition.
\newblock {\em arXiv preprint arXiv:1409.1556}, 2014.

\bibitem{stretcu2015multiple}
Otilia Stretcu and Marius Leordeanu.
\newblock Multiple frames matching for object discovery in video.
\newblock In {\em Proc. of BMVC}, 2015.

\bibitem{xiao2016track}
Fanyi Xiao and Yong~Jae Lee.
\newblock Track and segment: An iterative unsupervised approach for video
  object proposals.
\newblock In {\em Proc. of CVPR}, 2016.

\bibitem{yu2015fast}
Gang Yu and Junsong Yuan.
\newblock Fast action proposals for human action detection and search.
\newblock In {\em Proc. of CVPR}, 2015.

\bibitem{zitnick2014edge}
C~Lawrence Zitnick and Piotr Doll{\'a}r.
\newblock Edge boxes: Locating object proposals from edges.
\newblock In {\em Proc. of ECCV}. Springer, 2014.

\end{thebibliography}
}

\end{document}